\documentclass[lettersize,journal]{IEEEtran}
\usepackage{graphicx}
\usepackage{cite}
\usepackage{picinpar}
\usepackage{amsmath}
\usepackage{url}
\usepackage{flushend}
\usepackage[latin1]{inputenc}
\usepackage{colortbl}
\usepackage{soul}
\usepackage{multirow}
\usepackage{pifont}
\usepackage{color}
\usepackage{alltt}
\usepackage[hidelinks]{hyperref}
\usepackage{enumerate}
\usepackage{siunitx}
\usepackage{breakurl}
\usepackage{epstopdf}
\usepackage{pbox}
\usepackage{float}
\usepackage{amssymb}
\usepackage{xcolor}
\usepackage{booktabs}

\usepackage{algorithm}
\usepackage{algorithmic}

\begin{document}
\title{	Incremental Joint Learning of Depth, Pose and Implicit Scene Representation on Monocular Camera in Large-scale Scenes}

\author{
	Tianchen~Deng,
        Nailin~Wang,
        Chongdi~Wang,
        Shenghai~Yuan, 
         Jingchuan Wang,~\IEEEmembership{Senior Member,~IEEE}, \\
         Hesheng Wang,~\IEEEmembership{Senior Member,~IEEE},~Danwei~Wang,~\IEEEmembership{Life~Fellow,~IEEE},
        and~Weidong~Chen
\thanks{Tianchen Deng, Nailin Wang, Chongdi Wang, and Weidong Chen are with the Institute of Medical Robotics and Department
of Automation, Shanghai Jiao Tong University, and Key Laboratory of
System Control and Information Processing, Ministry of Education, Shanghai 200240, China. Danwei Wang and Shenghai Yuan are with the School of Electrical and Electronic Engineering, Nanyang Technological University, Singapore. This work is also supported by the National Key R\&D Program of China (Grant 2020YFC2007500), and the Science and Technology Commission of Shanghai Municipality (Grant 20DZ2220400). This research is supported by the National Research Foundation, Singapore, under the NRF Medium Sized Centre scheme (CARTIN), Maritime and Port Authority of Singapore under its Maritime Transformation Programme (Project No. SMI-2022-MTP-04), ASTAR under National Robotics Programme with Grant No.M22NBK0109. 
	}
}

\markboth{IEEE Transactions on Automation Science and Engineering}%
{Shell \MakeLowercase{\textit{et al.}}: A Sample Article Using IEEEtran.cls for IEEE Journals}

\maketitle
\begin{abstract}
Dense scene reconstruction for photo-realistic view synthesis has various applications, such as
VR/AR, and robotics navigation. Existing dense reconstruction methods are primarily designed for small room scenarios, but in practice, the scenes encountered by robots are typically large-scale environments. Most existing methods have difficulties in large-scale scenes due to three core challenges: \textit{(a) inaccurate depth input.} Depth information is crucial for both scene geometry reconstruction and pose estimation. Accurate depth input is impossible to get in real-world large-scale scenes. \textit{(b) inaccurate pose estimation.} Existing methods are not robust enough with the growth of cumulative errors in large
scenes and long sequences.  \textit{(c) insufficient scene representation capability.} A single global radiance field lacks the capacity to scale effectively to large-scale scenes. To this end, we propose an incremental joint learning framework, which can achieve accurate depth, pose estimation, and large-scale dense scene reconstruction. For depth estimation, a vision transformer-based network is adopted as the backbone to enhance performance in scale information estimation. For pose estimation, a feature-metric bundle adjustment (FBA) method is designed for accurate and robust camera tracking in large-scale scenes and eliminates pose drift. In terms of implicit scene representation, we propose an incremental scene representation method to construct the entire large-scale scene as multiple local radiance fields to enhance the scalability of 3D scene representation. In local radiance fields, we propose a tri-plane based scene representation method to further improve the accuracy and efficiency of scene reconstruction. We conduct extensive experiments on various datasets, including our own collected data, to demonstrate the effectiveness and accuracy of our method in depth estimation, pose estimation, and large-scale scene reconstruction. The code
has been open-sourced on \href{https://github.com/dtc111111/incre-dpsr}{https://github.com/dtc111111/incre-dpsr}.
\\
   
   \textit{Note to Practitioners}---In practical robot deployment scenarios, dense scene reconstruction can play a crucial role, such as in precise localization and obstacle avoidance. However, existing scene reconstruction methods are mainly suitable for small-room environments and struggle with larger-scale settings like long corridors or extended sequences. Our method introduces an incremental joint learning framework that simultaneously performs depth estimation, pose estimation, and scene reconstruction. Experimental results demonstrate the accuracy and effectiveness of this approach across different scene types.
\end{abstract}


\definecolor{limegreen}{rgb}{0.2, 0.8, 0.2}
\definecolor{forestgreen}{rgb}{0.13, 0.55, 0.13}
\definecolor{greenhtml}{rgb}{0.0, 0.5, 0.0}
\begin{figure}
    \centering
    \includegraphics[width=\linewidth]{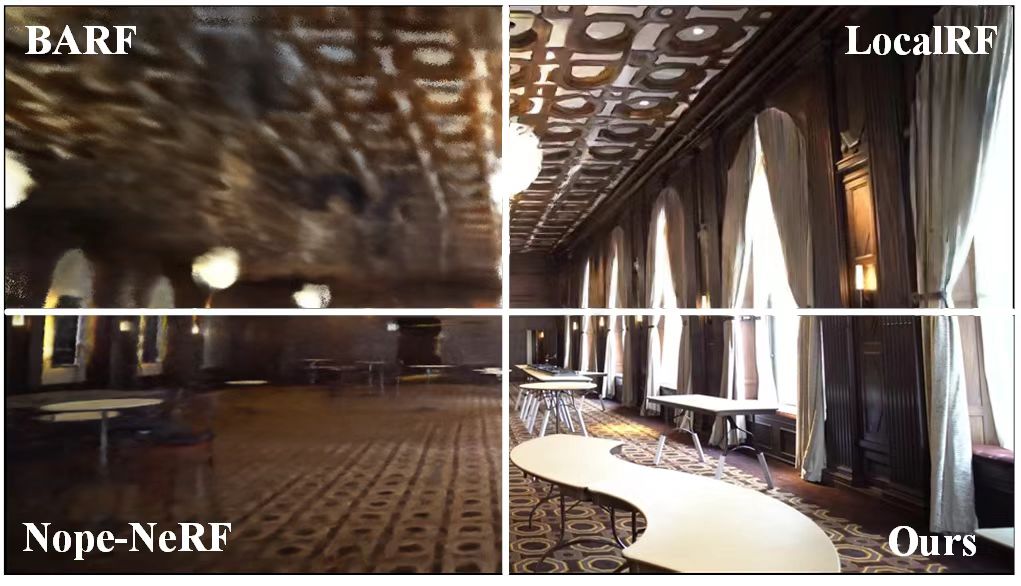}
    
    \caption{Qualitative results on the Tanks and Template datasets on Ballroom sequence. We present the visualization results of different methods' reconstruction results in the four corners of the image. Our method achieve better view synthesis performance than existing methods.  }
    
    \label{fig:reconstruct5}
    
\end{figure}
\section{Introduction}

\IEEEPARstart{I}{n} the past two decades, the demand for scene reconstruction has increased across various applications, including mobile robotics, autonomous driving. Many works focused on this field have made significant progress using Neural Radiance Fields (NeRF)~\cite{nerf}. However, the original NeRF method is only applicable to small room scenes and cannot perform accurate reconstruction in large-scale scenes, which limits the applicability of these methods in real-world robotic scenarios. In addition, the original NeRF method needs the accurate pose input, which makes it challenging to use in real-world deployments due to the inability to obtain accurate poses in practical robot operating environments. Pre-processing methods, such as Structure from motion (SFM), is also not robust and accurate in large-scale scenes. 

Some recent works are proposed to simultaneously estimate camera poses and neural implicit representation, such as BARF~\cite{barf}, SC-NeRF~\cite{scnerf}, and so on~\cite{plgslam,compact,neslam}.
Recent works, such as NoPe-NeRF~\cite{nopenerf} and LocalRF~\cite{progressive}, have been proposed to further improve the scene representation and pose estimation.
  NoPe-NeRF estimates the scale correction parameters during training to improve scene representation. LocalRF progressively estimates the poses of input frames to improve the scalability. However, both of them have difficulty in dealing with long trajectories and accumulating pose drifts. Their RGB loss-based pose estimation are also sensitive to pixel-wise noise variation in long trajectory sequences.


In this paper, we focus on large-scale 3D reconstruction without pose information, utilizing only monocular camera input. We highlight three key challenges in large-scale scenes:
\textit{(a) inaccurate depth scale estimation} Depth information is crucial for accurate geometry in neural scene representation,  particularly in large-scale scenes. However, existing methods neglect the importance of depth information in large-scale scene reconstruction and also fail to produce accurate depth.
\textit{(b) inaccurate pose estimation} Current methods struggle with pose estimation due to the error accumulation in large-scale scenes;
\textit{(c) limited scene representation capability} Existing approaches using a single, global model, which hinders scalability to larger scenes and longer sequences.

To this end, we propose an incremental joint learning framework for depth, pose estimation, and scene implicit representation in large-scale scenes. These three elements are closely interconnected: depth information is crucial for geometric representation of the scene. It can guide the neural point sampling and enhance scene geometry consistency. Accurate pose estimation is also vital for scene representation. So, we design a joint learning framework for these three elements. The depth estimation network adopts a vision transformer-based network to encode input images into high-level features. The transformer backbone utilizes a global receptive field to process features at every stage, which provides a more precise estimation.  
For the pose estimation network, a feature-metric bundle adjustment (FBA) method is proposed for robust pose estimation. Compared to the existing BA methods, our FBA method effectively avoids the sensitivity of pixel-wise noise variation, enabling robust localization in large-scale scenes. For scene implicit representation, existing methods usually use an MLP to encode the scene, leading to poor scene representation capability. We propose an incremental scene representation method. We dynamically initialize local radiance fields when the camera moves to the bound of the local scene representation. In local radiance fields, unlike the coordinate-based MLPs representation, we incorporate triplane with three corresponding axes XYZ. Our feature plane-based scene representation method effectively improves both the accuracy and efficiency of scene representation. Compared to the feature grid approach, our method substantially reduces the space complexity from cubic ($\mathcal{O}(n^3)$) to square ($\mathcal{O}(n)$). The incremental scene representation method can significantly enhance the scene representation capability in large-scale scenes. 

\textbf{The main contributions of this paper are as follows:}
\begin{itemize}
    \item A novel incremental joint learning framework is proposed for accurate depth estimation, pose estimation, and large-scale scene reconstruction.
    \item We design an feature-metric bundle adjustment method for the pose estimation module. Our method extracts multi-level features with pixel-wise confidences and incorporates a coarse-to-fine BA strategy to achieve robust and accurate pose estimation in large-scale scenes.
    \item An incremental scene representation method is designed, which dynamically initiates local radiance fields trained with frames within a local window. We also incorporate triplane based scene representation with three corresponding axes for accurate and efficient scene reconstruction. This enables accurate scene reconstruction in arbitrarily long video and large-scale scenes.
    \item Showcasing superior performance on both public datasets and proprietary elderly care sample tasks datasets, ensuring real-world effectiveness. The design and code will be made open-source to benefit the communities.

\end{itemize}

\section{Related work}
Depth estimation~\cite{monocular}, pose estimation~\cite{mglt,ctemlo,pointreg}, and scene representation based on monocular images have been a major problem in robotic perception and navigation~\cite{ppo}. Here, we summarize some relevant works.
\subsection{Joint Learning of Depth and Pose}
Depth and pose are closely interconnected components.
Some early works~\cite{deep} use groundtruth depth maps as supervision to learn monocular depth estimation from monocular images. \cite{synthetic} use synthetic images to help the disparity learning. Recently, some works have focused on
self-supervised learning to jointly estimate pose and depth. ~\cite{chen1,chen2,chen3,chen4} and \cite{scdepth} jointly optimize pose and depth  via photometric reprojection error. \cite{dual} focus on the scale ambiguity for monocular depth estimation with normalization.   However, convolutional neural networks do not have a global receptive field at every stage and fail to capture global consistency. 

\subsection{Implicit Scene Representation}
 With the proposal of NeRF~\cite{nerf}, many researchers explore combining this implicit scene representation into various applications, such as autonomous driving~\cite{prosgnerf,tase2,tw5,tw3}, and robot navigation. \cite{qiming1} integrates neural radiance fields with navigation networks end-to-end as a memory structure for the first time, highlighting the potential applications of neural radiance fields in downstream control domain tasks. \cite{qiming2} design a waypoint-based task decomposition strategy which effectively enhances the reliability of learning-based navigation. The key idea of NeRF is to model the radiance field of a scene with a multi-layer perceptron (MLP). However, the representation of volume densities can not commit the geometric consistency, leading to poor surface prediction for reconstruction tasks. UNISURF~\cite{unisurf} and NeuS~\cite{neus} are proposed to combine
world-centric 3D geometry representation with neural radiance fields. They replace the volume density with Signed
Distance Field (SDF) values, which can further improve the surface reconstruction.  
  \begin{figure*}
    \centering
    \includegraphics[width=\linewidth]{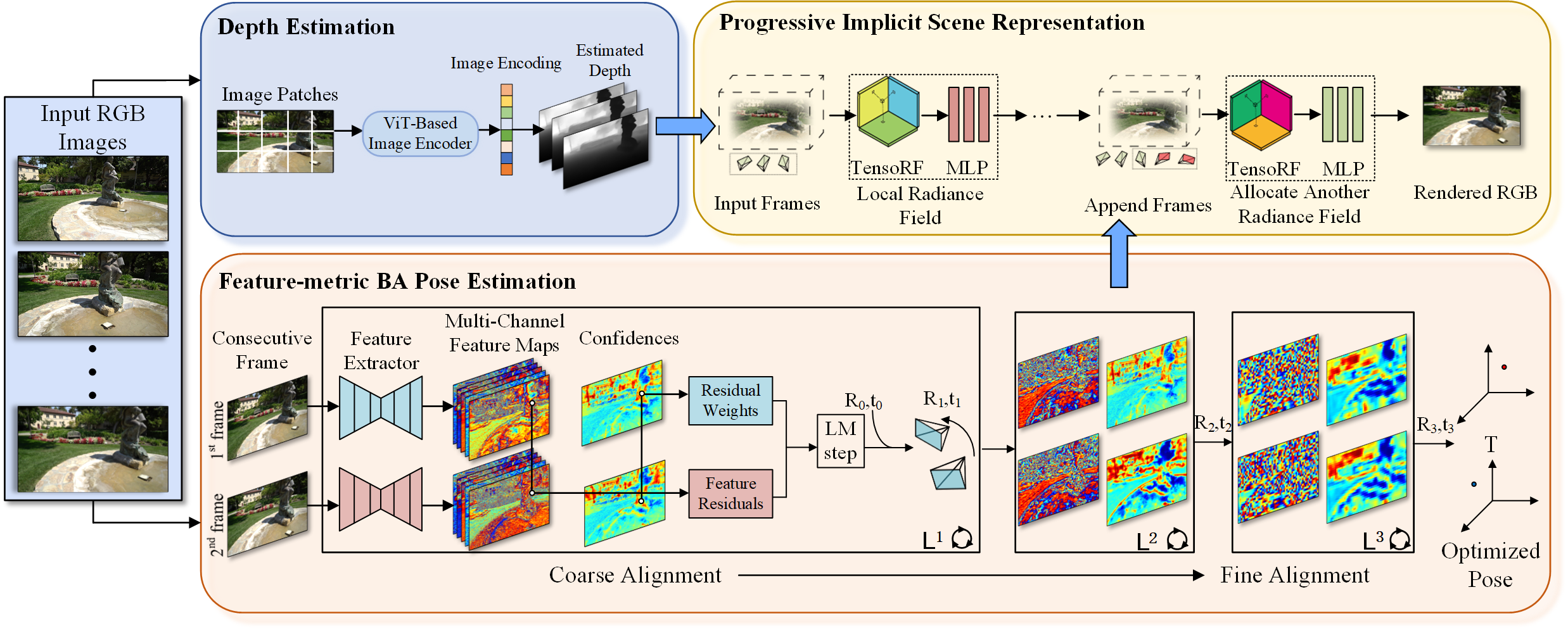}
    
    \caption{The pipeline of our system. There are three modules in our system: depth estimation module, Feature-metric Bundle Adjustment pose estimation module, and incremental scene implicit representation module. All three modules are jointly optimized during the system operation. }
    \label{fig:framework1}

\end{figure*}

\subsection{NeRF with Pose Optimization}
 The original NeRF method needs groundtruth pose information as their input, which is difficult to obtain in real-world environments.  They require pre-computed camera parameters
 obtained from SFM and SLAM algorithms~\cite{xie,xie2,liuu2,ram,tase3,tase4}.  Recently, many researchers have focused on implicit scene representation without pose prior.  iNeRF~\cite{inerf} is the first work focused on pose estimation and scene representation.  They estimate the camera pose for novel view images with a pre-trained NeRF model. NeRFmm~\cite{nerf--} jointly optimised as learnable parameters with NeRF training, through a
photometric reconstruction. BARF~\cite{barf} proposes a
 coarse-to-fine Bundle Adjustment method to estimate camera poses with neural 3D representation. \cite{zhu1,zhu2,li1,opengraph,openobj} combine semantic segementation methods~\cite{salt,sfpnet} with NeRF to improve the scene representation and camera tracking. Some other methods~\cite{densesplat,mneslam,mcnslam,vpgsslam} use multiple submaps to represent the scene. NoPe-NeRF~\cite{nopenerf} and LocalRF~\cite{progressive} are the most relevant work to ours. NoPe-NeRF learns scale and shift parameters during training and incorporates undistorted monocular depth prior. However, they often fail in large-scale scene representation due to the inaccurate depth estimation, the accumulated pose drift, and the limited scene representation capacity. Their pose estimation module are not robust and accurate in large-scale scenes, resulting in increased pose drifts.
 
 \section{Joint learning of depth, pose, and implicit scene representation}
 In this work, we propose a joint learning framework for depth, pose estimation, and implicit scene representation. The architecture of our framework is shown in Fig.~\ref{fig:framework1}. The system only takes in monocular RGB images $I\in H \times W \times3$. There are three networks in our system: depth estimation network, pose estimation network, and scene representation network. We will introduce our three networks separately.

 \subsection{Depth Estimation}
 In our depth network, we choose the vision transformer network as our backbone. Vision transformer network~\cite{vit} has shown strong feature learning capabilities. Most large vision models, such as Segment Anything Model \cite{sam}, choose it as their backbone. Inspired by this, we leverage vision transformers in place of convolutional neural networks for dense depth prediction. We maintain an encoder-decoder structure for our depth network. For the input images $I\in H \times W \times3$, these images are decomposed into a sequence of 2D patches, where $(H, W)$ is the height and width of the original image, $(P, P)$ is the resolution of each image patches. We use ResNet50 with group norm and weight standardization to encode these image patches. Our vision transformer module transforms these patches using sequential blocks of multi-headed self-attention (MSA).
\begin{equation}
\begin{aligned}
& \hat{Z}^l=MSA\left(L N\left(Z^{l-1}\right)\right)+Z^{l-1} \\
& Z^l=M L P\left(L N\left(\hat{Z}^l\right)\right)+\hat{Z}^l \\
& \hat{Z}^{l+1}=MSA\left(L N\left(Z^l\right)\right)+Z^l \\
& Z^{l+1}=M L P\left(L N\left(\hat{Z}^{l+1}\right)\right)+\hat{Z}^{l+1},
\end{aligned}
\end{equation}
where $Z^{l+1}$ is the output feature in stage $l$. LN represents layer norm operation. Then, we adopt a depth estimation head with a convolution layer and linear layer to decode the feature. The final linear layer projects this representation to a non-negative scalar that represents the inverse depth prediction for every pixel. Bilinear interpolation is used to upsample the representation. This depth network can provide dense and accurate depth predictions in large-scale scenes.

\subsection{Feature-metric BA Pose Estimation}
In terms of pose estimation, we propose a Feature-metric bundle adjustment network. We show the architecture of our network in Fig.~\ref{fig:framework1}. Inspired by~\cite{pixloc}, each image is encoded into high-level features and uses FBA to get an accurate and robust pose estimation. For every consecutive frames $I_a,I_b$, we use a Unet-like encoder-decoder network to extract $D_l$ dimensional feature map $\mathbf{F}^l \in \mathbb{R}^{W_l \times H_l \times D_l}$ at each level $l \in\{L, \ldots, 1\}$. Those multi-channel image features with decreasing resolution can encode richer semantic information and larger spatial information of the image. Our network can also simultaneously estimate the confidence of each pixel to select the robust features for pose estimation. 
Compared with the original BA method, this learned representation is robust to large illumination or viewpoint changes, which can provide accurate pose estimation in indoor and outdoor environments. Despite poor initial pose estimates, it can also provide meaningful gradients for successful estimation. Those joint learning methods with classical direct alignment \cite{nopenerf,progressive} operate on the original RGB image, which is not robust to large-scale scenes and long-term changes, and resorts to Gaussian image pyramids, which still largely limits the convergence to frame-to-frame tracking.

\noindent \textbf{Feature Residual:}
We define the direct feature residual between the two consecutive frames. We optimize the relative pose$\{\mathbf{R},\mathbf{t}\}$ through minimizing the feature residual. For a given feature level $l$ and each 3D point $i$ observed in two consecutive frames, we define a residual:
\begin{equation}
\mathbf{r}_a^i=\mathbf{F}_b^l\left[\Pi\left(\mathbf{R} \mathbf{P}_i+\mathbf{t}\right)_b\right]-\mathbf{F}_a^l\left[\mathbf{p}_a^i\right]
\end{equation}
where $\mathbf{p}_a^i=\Pi\left(\mathbf{R} \mathbf{P}_i+\mathbf{t}\right)_b$ is the projection of frame b given its current pose estimate. $[\cdot]$ is a lookup with sub-pixel interpolation. $F_a^l,F_b^l$ are the corresponding feature maps of images a and b. The total error over N observations is:
\begin{equation}
E_l(\mathbf{R}, \mathbf{t})=\sum_{i, a} w_a^i \rho\left(\left\|\mathbf{r}_a^i\right\|_2^2\right)
\end{equation}

where $\rho(\cdot)$ is a robust cost function with derivative $\rho'$,
and $\omega_a^i$ is a per-residual weight. We use the Levenberg-Marquardt (LM) algorithm to iteratively optimize this nonlinear least-squares cost from an initial estimate. Follow \cite{pixloc}, each pose update is parametrized on the SE(3) manifold using its Lie algebra. 
\begin{equation}
\boldsymbol{\delta}=-(\mathbf{H}+\lambda \operatorname{diag}(\mathbf{H}))^{-1} \mathbf{J}^{\top} \mathbf{W r}
\end{equation}
where $\lambda$ is the damping factor and H is the Hessian matrices. Then, the final pose can be defined by left-multiplication on the manifold SE(3).
\begin{algorithm}[t]
	\renewcommand{\algorithmicrequire}{\textbf{Input:}}
	\renewcommand{\algorithmicensure}{\textbf{Output:}}
	\renewcommand{\algorithmicendwhile}{\textbf{end}}
	\caption{Incremental Implicit Scene Representation.}
	\label{alg1}
	\begin{algorithmic}[1]
		\STATE $j\gets 1$;  \hfill $/\ast$ The first scene representation index$\ast/$
		\STATE $p\gets 1$; \hfill $/\ast$ First camera pose index, P denotes the length of sequence$\ast/$
		\STATE $q\gets 1$; \hfill $/\ast$Start with the first frames $\ast/$
		\STATE $\{R,T\}_{p..q} \gets 1$; \hfill $/\ast$ Initialize poses as identity$\ast/$
		\STATE $\theta_j \gets InitializeLR()$; \hfill $/\ast$ Initialize the first local radiance fields$\ast/$
		\WHILE{$q<P$}
        \STATE $m \gets 0$
		\WHILE{$m<100$}
		\STATE Optimize $(\theta_j)$; \hfill $/\ast$ Optimize the first local radiance field with the first frame$\ast/$
		\STATE $m\gets m+1$;
		\ENDWHILE
		
		\WHILE{$\{R,t\}_j< Bound_j$} 
		\STATE $q\gets q+1$;
		\STATE $\{R,T\}_{q}\gets \{R,T\}_{q-1}$; \hfill $/\ast$ Append a pair of pose R,T at the end of the trajectory $\ast/$
		\STATE  Optimize($\{R,T\}_{p..q},\theta_j$); \hfill $/\ast$ Refine poses and local radiance field $\ast/$
		\ENDWHILE
        \IF{$q<P$}
        \STATE $j\gets j+1$; \hfill $/\ast$ A new local radiance field index$\ast/$
        \STATE $\theta_j\gets InitializeLR()$ \hfill $/\ast$ Initialize a new local radiance field$\ast/$
        \STATE $t_j \gets t_q$;  \hfill $/\ast$ Centered around the last pose$\ast/$
        \STATE $p \gets p+n$;  \hfill $/\ast$ Stop considering the frames in last local radiance field$\ast/$
		\ENDIF
		\STATE Repeat until all frames are registered.
		\ENDWHILE
        
	\end{algorithmic}  
\label{alg:incremental}
\end{algorithm}


\noindent\textbf{Feature-metric BA loss:} Our network is trained by minimizing the FBA of the corresponding pixel from coarse level to fine level:
\begin{equation}
\mathcal{L}_{FBA}=\frac{1}{L} \sum_l \sum_i\left\|\Pi\left(\mathbf{R}_l \mathbf{P}_i+\mathbf{t}_l\right)-\Pi\left(\overline{\mathbf{R}} \mathbf{P}_i+\overline{\mathbf{t}}\right)\right\|_\gamma
\end{equation}
where $\gamma$ is the Huber cost, L denotes the different feature level. With coarse-to-fine pose estimation, low-resolution feature maps are responsible for the robustness of the pose estimation, while finer features enhance its accuracy. This feature-metric BA loss weights the supervision of the rotation and translation adaptively for each training example. It greatly enhances the accuracy and robustness of our pose network in large-scale scenes. 

\begin{figure}[t]
    \centering
    \includegraphics[width=\linewidth]{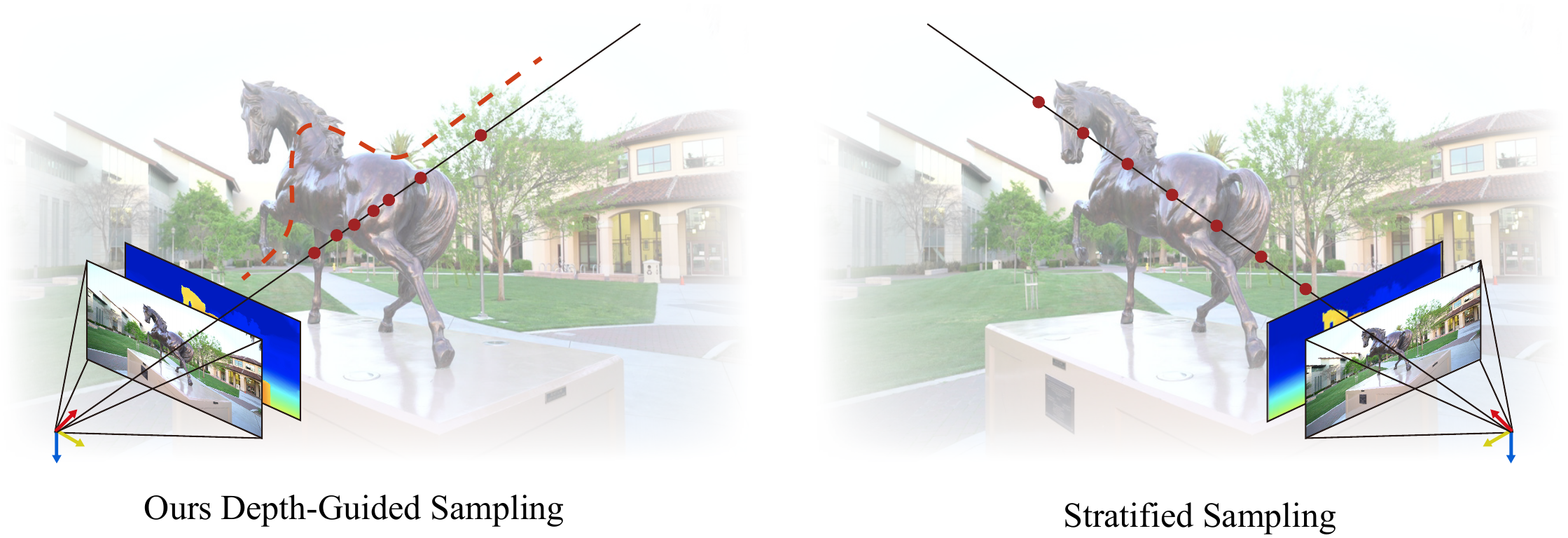}
    \vspace{-0.3cm}
    \caption{We illustrate the differences between our sampling method and original sampling method. Our method makes better use of the estimated depth information to guide the ray sampling process. }
    \label{fig:sample}
    \vspace{-0.3cm}
\end{figure}
\begin{figure*}
    \centering
    
    \includegraphics[width=\linewidth]{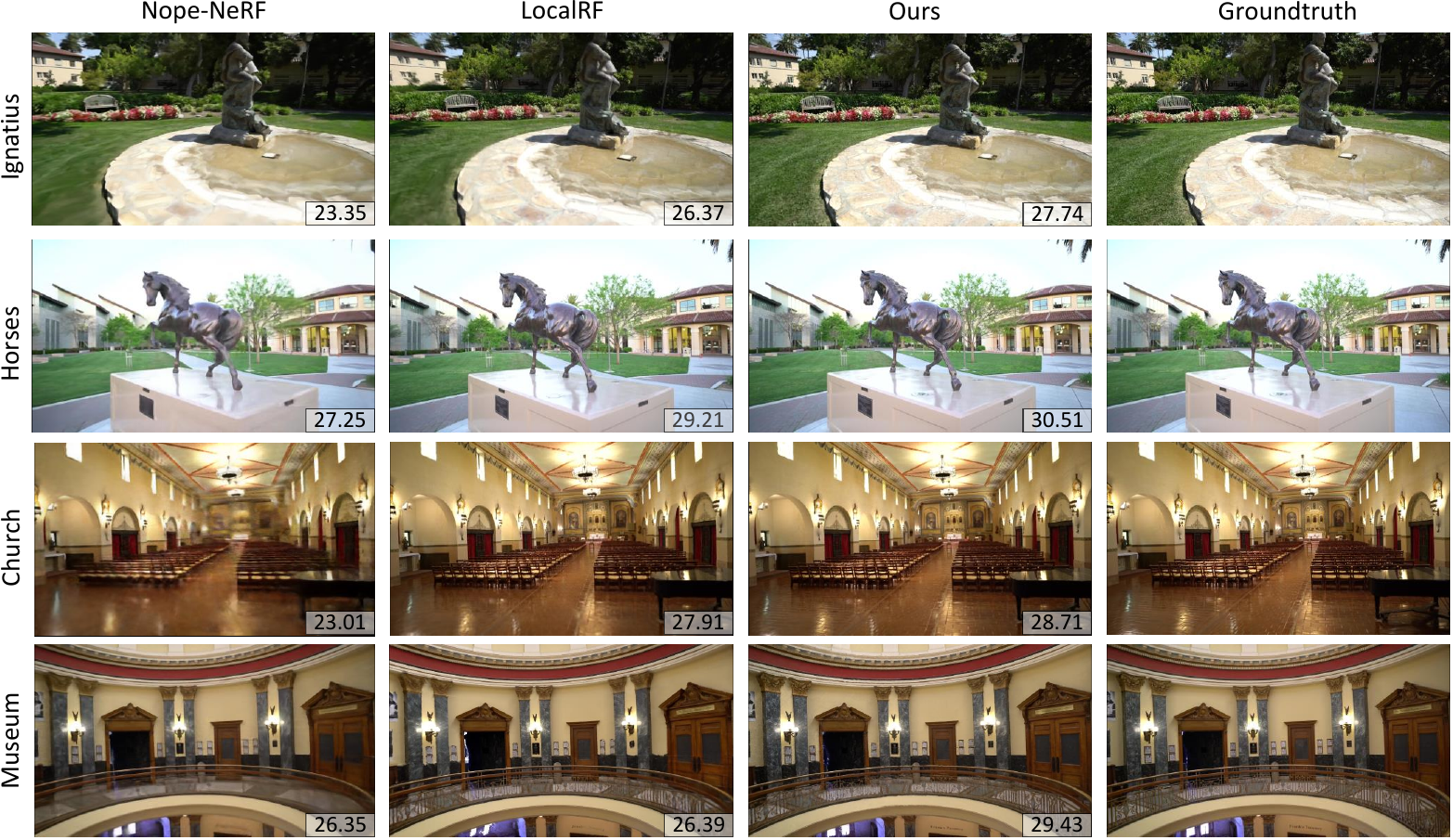}
    
    \caption{Qualitative results on Tanks and Temples datasets~\cite{tanks}. We present the scene reconstruction results of NoPe-NeRF\cite{nopenerf}, LocalRF~\cite{progressive}, and our method with PSNR value. Our method performs better than other methods with higher-quality view synthesis results.}
    \label{fig:reconstruct}

\end{figure*}
\subsection{Incremental Implicit Scene Representation}

Existing scene implicit scene representation methods such as \cite{scnerf,barf} can only achieve satisfactory results in small-scale scenes. They are frequently constrained by their limitations in scene representation capabilities when dealing with longer sequences and tend to converge towards local minima during the optimization process. Inspired by~\cite{progressive}, we propose an incremental implicit scene representation. The architecture of our network is shown in Fig.~\ref{fig:framework1} and Alg.~\ref{alg:incremental}. At the beginning, we will initialize a local neural radiance field and optimize it using the initial few frames as input. As the robot moves forward, we progressively append new frames within the local neural radiance field and iteratively refine the scene representation, pose estimation, and depth estimation. This procedure persists until we reach the boundary of the local neural radiance field. Then, we dynamically allocate new local radiance fields throughout the optimization with frames within a sliding window. This procedure will continue until all the images are registered into the network. So, the entire scene can be represented as multiple local NeRF: 
\begin{equation}
\{I_i,D_i\}_{i=1}^M\mapsto \{\mathrm{LR}^1_{\theta_1},\mathrm{LR}^2_{\theta_2},\dots ,\mathrm{LR}^n_{\theta_n}\}\mapsto\{\mathbf{c},\mathbf{\sigma}\}
\end{equation}
where $D_i$ is the estimated depth image from our depth network, $\mathbf{c}$ denotes the image color, $\sigma$ represents the volume density.
Whenever the estimated camera pose exceeds the boundaries of the current local radiance fields, we cease optimizing previous poses and freeze the network parameters. This allows us to decrease memory usage by discarding supervisory frames that are no longer needed.
During the optimization, we jointly optimize the depth, pose, and scene representation in the local radiance fields. 
\renewcommand\arraystretch{1.3}
\begin{table*}[]
\centering
\caption{Novel view synthesis results on Tanks and Temples~\cite{tanks}. We use PSNR, SSIM, and LPIPS as our metrics. }
\scalebox{0.9}{
\setlength{\tabcolsep}{1mm}{
\begin{tabular}{llcccccccccccccccccc}
\hline
\multicolumn{1}{c}{\multirow{2}{*}{scenes}} & \multirow{2}{*}{} & \multicolumn{3}{c}{BARF\cite{barf}} & \multicolumn{3}{c}{NoPe-NeRF~\cite{nopenerf}} & \multicolumn{3}{c}{LocalRF~\cite{progressive}} & \multicolumn{3}{c}{Flow-NeRF~\cite{flownerf}} & \multicolumn{3}{c}{RA-NeRF~\cite{ranerf}} & \multicolumn{3}{c}{Ours}                       \\ \cline{3-20} 
\multicolumn{1}{c}{}                        &                   & PSNR       & SSIM      & LPIPS      & PSNR         & SSIM        & LPIPS       & PSNR  & SSIM          & LPIPS          & PSNR      & SSIM              & LPIPS    & PSNR              & SSIM    & LPIPS    & PSNR           & SSIM          & LPIPS         \\ \hline
\multirow{9}{*}{\rotatebox{90}{\textbf{Tanks and Temples}}}          & Ballroom          & 20.66      & 0.50      & 0.30       & 24.97        & 0.71        & 0.20        & 25.29 & 0.89          & \textbf{0.08}  & 28.83     & \textbf{0.86}     & 0.24     & \textbf{30.25}    & N/A     & 0.11     & 26.50          & 0.85          & \textbf{0.08} \\
                                            & Barn              & 25.28      & 0.64      & 0.24       & 25.77        & 0.68        & 0.22        & 27.89 & 0.88          & \textbf{0.11}  & 28.53     & 0.78              & 0.35     & 25.99             & N/A     & 0.38     & \textbf{28.59} & \textbf{0.91} & \textbf{0.11} \\
                                            & Church            & 23.17      & 0.62      & 0.27       & 23.60        & 0.67        & 0.23        & 28.07 & 0.91          & \textbf{0.07}  & 28.27     & 0.83              & 0.28     & \textbf{32.05}    & N/A     & 0.08     & 28.82          & \textbf{0.92} & 0.08          \\
                                            & Family            & 23.04      & 0.61      & 0.28       & 23.77        & 0.68        & 0.22        & 28.60 & 0.92          & \textbf{0.007} & 29.40     & 0.85              & 0.29     & 30.46             & N/A     & 0.15     & \textbf{30.48} & \textbf{0.94} & 0.05          \\
                                            & Francis           & 25.85      & 0.69      & 0.28       & 29.48        & 0.80        & 0.18        & 31.71 & \textbf{0.93} & 0.11           & 30.63     & 0.83              & 0.33     & 32.04             & N/A     & 0.24     & \textbf{33.38} & \textbf{0.93} & \textbf{0.10} \\
                                            & Horse             & 24.09      & 0.72      & 0.21       & 27.30        & 0.83        & 0.13        & 29.98 & \textbf{0.94} & \textbf{0.07}  & 28.57     & 0.86              & 0.23     & 28.26             & N/A     & 0.18     & \textbf{30.07} & 0.93          & \textbf{0.07} \\
                                            & Ignatius          & 21.78      & 0.47      & 0.30       & 23.77        & 0.61        & 0.23        & 26.54 & 0.86          & \textbf{0.09}  & 26.25     & 0.73              & 0.35     & 27.77    & N/A     & 0.25     & \textbf{27.83}          & \textbf{0.87} & \textbf{0.09} \\
                                            & Museum            & 23.58      & 0.61      & 0.28       & 26.39        & 0.76        & 0.17        & 26.35 & 0.86          & 0.16           & 29.43     & 0.85              & 0.27     & \textbf{31.59}    & N/A     & 0.10     & 29.43          & \textbf{0.92} & \textbf{0.12} \\
                                            & mean              & 23.42      & 0.61      & 0.27       & 26.34        & 0.74        & 0.19        & 27.87 & \textbf{0.91} & \textbf{0.09}  & 28.73     & 0.82              & 0.29     & 29.80    & N/A     & 0.18     & \textbf{29.84}          & 0.90          & \textbf{0.09} \\ \hline
\end{tabular}}
}
\label{tab:view}
\end{table*}

\noindent \textbf{Local Scene Representation}
Voxel grid-based architectures~\cite{niceslam} are the mainstream in NeRF-based systems. However, they struggle with cubical memory growing and space complexity. In our local scene representation, we factorize the local radiance fields to compact components and reduce the space complexity from cubic ($\mathcal{O}(n^3)$) to square ($\mathcal{O}(n)$). The 3D feature grid $\mathcal{G}$ can be factorized as with Vector-Matrix (VM) decomposition:
\begin{equation}
    \mathcal{G}=\sum_{r=1}^{R} \mathbf{v}_{r}^X \circ \mathbf{M}_{r}^{Y Z}+\mathbf{v}_{r}^Y \circ \mathbf{M}_{r}^{X Z}+\mathbf{v}_{r}^Z \circ \mathbf{M}_{r}^{X Y}
\end{equation}
where $\mathbf{v}_{r}^X$,$\mathbf{v}_{r}^Y$, $\mathbf{v}_{r}^Z$ are the vectors corresponding to XYZ axis. Our factorization-based model efficiently computes the feature vector for each voxel at a low cost, requiring only a single value per XYZ-mode vector or matrix factor. Additionally, we incorporate a tri-linear interpolation method to achieve a continuous fields. In our local neural radiance field, we can just interpolate the vector/matirx factor for the corresponding modes. Instead of recovering eight individual tensor elements for trilinear interpolation, we directly compute the interpolated value, significantly reducing computation and memory costs during runtime.

\noindent \textbf{Differentiable Rendering}
For every input image $I$, we use the camera intrinsic matrix and camera pose to generate rays $r=x_i+td_i$. We sample points $x_i$ along rays. Then, we query our local radiance fields to get the color and density. $F_{\Theta}:(\mathbf{x_i}, \mathbf{d_i}) \rightarrow(\mathbf{c}, \sigma)$. Through differentiable rendering, we can render along the rays: 
\begin{equation}
\begin{aligned}
\hat{\mathbf{C}}(\mathbf{r}) & =\sum_{i=1}^N T_i\left(1-\exp \left(-\sigma_i \delta_i\right)\right) \mathbf{c}_i, \\
T_i & =\exp \left(-\sum_{i-1}^j \sigma_j \delta_j\right) 
\end{aligned}
\end{equation}
where $\delta_i$ is the distance between sample points, $T_i$
represents the accumulated transmittance along the ray, and $N$ denotes the number of sample points along the ray. Compared with existing scene representation methods, we have re-parameterized the scene representation for unbounded scenes similar to \cite{mipnerf360}. We use a contract function to map every point into a cubic space:
\begin{equation}
\operatorname{contract}(\mathbf{x})= \begin{cases}\mathbf{x} & \text { if }\|\mathbf{x}\|_{\infty} \leq 1 \\ \left(2-\frac{1}{\|\mathbf{x}\|_{\infty}}\right)\left(\frac{\mathbf{x}}{\|\mathbf{x}\|_{\infty}}\right) & \text { otherwise. }\end{cases}
\end{equation}
This contract function can help our method to reconstruct unbounded scenes.

\noindent \textbf{Depth Guided Sampling} \quad
The estimated depth images provide valuable geometry information which can guide neural point sampling along a ray. We get $N_{strat}$ points for stratified sampling between the near and far planes. Then, $N_{surface}$ points are drawn from the Gaussian distribution determined by the depth prior. When the depth is not known or invalid, we use the rendered depth. Compared to the original methods, our approach allows for more efficient point sampling and enhances the scene representation capability of the network.  We illustrate the differences between our sampling method and original sampling method in Fig.~\ref{fig:sample}.

\begin{figure*}
    \centering
    \includegraphics[width=\linewidth]{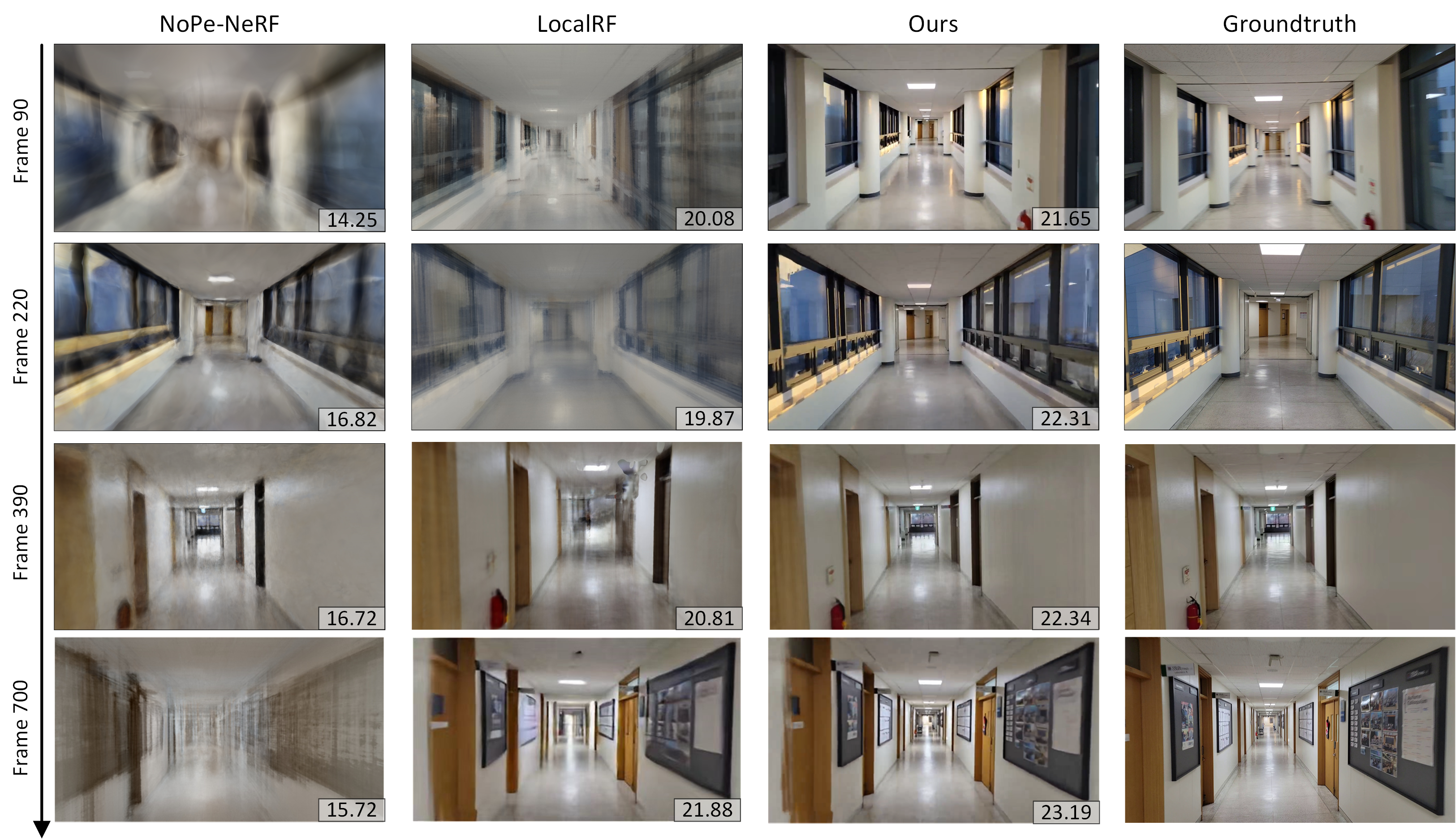}
    
    \caption{Qualitative results on Static Hikes Indoor datasets~\cite{progressive} with long trajectory and span over 1000 images. We present the scene reconstruction results of NoPe-NeRF\cite{nopenerf}, LocalRF~\cite{progressive}, and our method in different time steps (90, 220, 390, 700). Our incremental scene representation method significantly improves the scene reconstruction performance in large-scale scenes. }
    \label{fig:indoor}
    
\end{figure*}

\subsection{Incremental Joint Learning}
In this section, we provide more details of the optimization
of depth $D$, RGB $C$, camera poses $\{R, t\}$, and scene representation $\theta$. In order to optimize our depth network and render depth images, we propose depth loss and estimated depth function.
\begin{equation}
\hat{\mathbf{D}}(\mathbf{r})=\sum_{i=1}^N T_i\left(1-\exp \left(-\sigma_i \delta_i\right)\right) d_i
\end{equation}
\begin{equation}
\mathcal{L}_d=\left|\hat{\mathbf{D}}^*-\mathbf{D}^*\right|
\end{equation}
where $\hat{D}^{*}(r)$ is the normalized render depth, $D^*(r)$ is the normalized estimated depth image from our depth network. We use a pre-trained model to get depth images prior to our depth network. Since the monocular depth is not scale- and shift-invariant, we normalize the depth and estimate the scale and shift of depth:
\begin{equation}
    t(D)=\frac{1}{N}\sum_{i=1}^{N}(D)
\end{equation}
\begin{equation}
    s(D)=\frac{1}{N}\sum_{i=1}^{N}|D-t(D)|
\end{equation}
We sample N pixels for the frame. Then, we normalize the depth:
\begin{equation}
    D^*=\frac{D-t(D)}{s(D)}
\end{equation}
We define a photo-metric loss to optimize local radiance fields:
\begin{equation}
\mathcal{L}_p=\|\hat{\mathbf{C}}(\mathbf{r})-\mathbf{C}(\mathbf{r})\|_2^2
\end{equation}
where $\hat{C}(r)$ is the render image and $C(r)$ is the input RGB image. We also add an optical flow loss
between neighbor frames as they have proven to improve the accuracy and robustness in challenging scenarios. 
\begin{equation}
\hat{\mathcal{F}}_{k \rightarrow k+1}=(u, v)-\Pi\left(\{R,t\}_{k \rightarrow k+1} \Pi^{-1}(u, v, \hat{D})\right)
\end{equation}
\begin{table*}[]
\caption{Pose estimation results on Tanks and Temples datasets~\cite{tanks}. We use ATE and RPE as our metrics.}
\centering
\begin{tabular}{cccccccccccccc}
\toprule
\multicolumn{2}{c}{\multirow{2}{*}{Scenes}}   & \multicolumn{3}{c}{BARF~\cite{barf}} & \multicolumn{3}{c}{NoPe-NeRF~\cite{nopenerf}} & \multicolumn{3}{c}{LocalRF~\cite{progressive}} & \multicolumn{3}{c}{Ours} \\ \cline{3-14} 
\multicolumn{2}{c}{}                          & ATE$\downarrow$     & RPE$_r$$\downarrow$   & RPE$_t$$\downarrow$   & ATE$\downarrow$      & RPE$_r$$\downarrow$    & RPE$_t$$\downarrow$   & ATE$\downarrow$     & RPE$_r$$\downarrow$   & RPE$_t$$\downarrow$   & ATE$\downarrow$   & RPE$_r$$\downarrow$   & RPE$_t$$\downarrow$  \\ \midrule
\multirow{8}{*}{\rotatebox{90}{Tanks and Temples~\cite{tanks}}} & Ballroom & 0.019  & 0.228  & 0.343  & 0.003          & 0.026          & 0.054          & \textbf{0.014} & 0.077 & \textbf{0.052} & \textbf{0.014}          & \textbf{0.022} & \textbf{0.051} \\
                                   & Barn     & 0.075  & 0.326  & 1.402  & 0.023          & 0.034          & 0.127          & 0.019          & 0.088 & 0.048          & \textbf{0.019} & \textbf{0.015} & \textbf{0.044} \\
                                   & Church   & 0.059  & 0.063  & 0.458  & 0.026          & 0.013          & 0.088          & 0.017          & 0.077 & \textbf{0.042} & \textbf{0.016} & \textbf{0.023} & 0.111          \\
                                   & Family   & 0.116  & 0.595  & 0.577  & 0.006          & 0.045          & 0.049          & \textbf{0.004} & 0.074 & 0.043          & \textbf{0.004}          & \textbf{0.021} & \textbf{0.035} \\
                                   & Francis  & 0.095  & 0.749  & 0.924  & 0.005          & 0.029          & 0.069          & \textbf{0.004}          & 0.035 & 0.082          & \textbf{0.003} & \textbf{0.025} & \textbf{0.068} \\
                                   & Horse    & 0.016  & 0.399  & 0.239  & \textbf{0.005} & 0.032          & 0.192          & 0.037          & 0.287 & 0.561          & 0.014          & \textbf{0.027} & \textbf{0.035} \\
                                   & Ignatius & 0.057  & 0.288  & 1.187  & 0.003          & \textbf{0.010} & 0.039          & 0.003 & 0.076 & \textbf{0.028} & \textbf{0.002}          & 0.017          & 0.061          \\
                                   & Museum   & 0.257  & 1.128  & 2.589  & 0.035          & 0.247          & 0.335          & \textbf{0.025} & 0.303 & \textbf{0.179} & \textbf{0.025}          & \textbf{0.219} & 0.217          \\
                                   & Mean     & 0.087  & 0.472  & 0.965  & \textbf{0.006} & 0.056          & \textbf{0.080} & 0.015          & 0.101 & 0.082          & 0.012          & \textbf{0.046} & \textbf{0.077} \\ \bottomrule
\end{tabular}
\label{tab:pose}

\end{table*}

Where $\Pi^{-1}$ represents the back-projection from pixel coordinates into 3D points. $\{R,t\}_{k \rightarrow k+1}$ means the relative camera pose transformation. Then, we calculate our optical flow loss:
\begin{equation}
\mathcal{L}_{fa}=\left\|\hat{\mathcal{F}}_{k \rightarrow k+1}-\mathcal{F}_{k \rightarrow k+1}\right\|_1
\end{equation}
This is the forward optical flow, and we also calculate backward optical flow $L_{fb}$ to supervise our implicit scene representation network.
The final loss of our network is :
\begin{equation}
    \mathcal{L}=\lambda_1\mathcal{L}_p+\lambda_2\mathcal{L}_D+\lambda_3\mathcal{L}_{FBA}+\lambda_4\mathcal{L}_{fa}+\lambda_5\mathcal{L}_{fb}
\end{equation}
We use the progressive optimization scheme to improve the robustness and to satisfy large-scale scenes and long sequences. This joint optimization framework simultaneously leverages 2D photometric consistency from RGB images and 3D structural information from depth, enforcing coherence between 2D appearance and 3D geometry. This synergy leads to enhanced accuracy in depth estimation, pose prediction, and overall scene reconstruction. We start by using a small amount of frames (for example five frames) to initiate the operation. Subsequently, we progressively introduce new frames into the optimization process in a sequential manner.

\begin{table}[]
\caption{The depth estimation accuracy on the Tanks and Temples dataset~\cite{tanks} and indoor datasets~\cite{progressive}, using Depth L1, RMSE, Absolute Relative Error (Abs Rel), and Square Relative Error (Sq Rel) as metrics.}
\centering
\setlength{\tabcolsep}{0.8mm}{
\begin{tabular}{ccccc}
\cline{1-5}
Scenes                         &                      & NoPe-NeRF~\cite{nopenerf}            & LocalRF~\cite{progressive}              & Ours                 \\ \cline{1-5}
\multirow{4}{*}{Tanks/Francis~\cite{tanks}} & Abs Rel$\downarrow$              & 0.658                & 0.690                & \textbf{0.657}         \\
                               & Sq Rel$\downarrow$               & 135.46               & 140.96               & \textbf{135.42}        \\
                               & Depth-L1 [cm] $\downarrow$             & 159.57               & 176.79               & \textbf{158.77}      \\
                               & RMSE [cm] $\downarrow$                 & \textbf{171.50}      & 191.45               & 171.69               \\ \cline{1-5}
\multirow{4}{*}{Hike/Indoor~\cite{progressive}}   & Abs Rel$\downarrow$              & 0.734                & 0.617                & \textbf{0.544}       \\
                               & Sq Rel$\downarrow$               & 140.91               & 103.45               & \textbf{88.73}        \\
                               & Depth-L1$\downarrow$             & 187.27               & 153.88               & \textbf{136.2}       \\
                               & RMSE$\downarrow$                 & 189.55               & 160.22               & \textbf{148.49}      \\ \hline

\end{tabular}}
\label{tab:depth}
\end{table}

\begin{table}[]
\centering
\caption{The scene reconstruction results on Tanks and temples dataset. We report the average PSNR, SSIM, LPIPS results.}
\begin{tabular}{lllll}
\toprule
\multicolumn{2}{l}{Methods}                   & PSNR$\uparrow$           & SSIM$\uparrow$          & LPIPS$\downarrow$         \\ \midrule
\multirow{5}{*}{COLMAP}         & NeRFfacto\cite{nerfstudio}   & 24.17          & 0.65          & 0.194          \\
                                & Mip-NeRF360\cite{mipnerf360} & 27.74          & 0.77          & 0.152          \\
                                & NoPe-NeRF\cite{nopenerf}   & 26.43          & 0.74          & 0.184          \\
                                & LocalRF\cite{progressive}     & 28.04          & 0.88          & 0.096          \\
                                & Ours        & \textbf{28.17} & \textbf{0.88} & \textbf{0.095} \\ \midrule
\multirow{6}{*}{Joint learning} & SCNeRF \cite{scnerf}     & 20.82          & 0.58          & 0.227          \\
                                & NeRFmm \cite{nerf--}     & 22.53          & 0.64          & 0.297          \\
                                & BARF \cite{barf}       & 23.42          & 0.64          & 0.275          \\
                                & NoPe-NeRF \cite{nopenerf}  & 26.34          & 0.74          & 0.194          \\
                                & LocalRF \cite{progressive}    & 27.87          & 0.91          & 0.094          \\
                                & Ours        & \textbf{29.04} & \textbf{0.90} & \textbf{0.092} \\ \bottomrule
\end{tabular}

\label{tab:tanks}
\end{table}

\section{Experiments}
In this section, we present our experimental results on different real-world datasets and conduct a comprehensive ablation study to verify the effectiveness of our method. We also gather proprietary real-world datasets using a robotic wheelchair to assess performance on mobile platforms in real-world applications.

\noindent \textbf{Baselines Methods and Public Datasets.} The main baseline methods we compared are BARF~\cite{barf}, NoPe-NeRF~\cite{nopenerf},  LocalRF~\cite{progressive}, Flow-NeRF~\cite{flownerf}, RA-NeRF~\cite{ranerf}. All of them are joint learning methods of camera poses and implicit scene representation. We also compare our method against SCNeRF~\cite{scnerf}, NeRFmm~\cite{nerf--}, NeRFfacto~\cite{nerfstudio}. We evaluate our method on the
Tanks and Temples dataset~\cite{tanks} (from 4x4m to 8x8m) and Static Hikes dataset~\cite{progressive} (15x15m). 
The Tanks and Temples dataset features slow-moving image sequences spanning a few meters in both indoor and outdoor scenes. 
We also utilize the Static Hikes indoor dataset, which includes hand-held sequences with longer camera trajectories, testing scalability and pose estimation robustness over ranges typically exceeding 100 meters (30x30m). We strictly follow the experimental settings of our baseline to partition the training set and the test set.

\begin{figure}
    \centering
    \includegraphics[width=\linewidth]{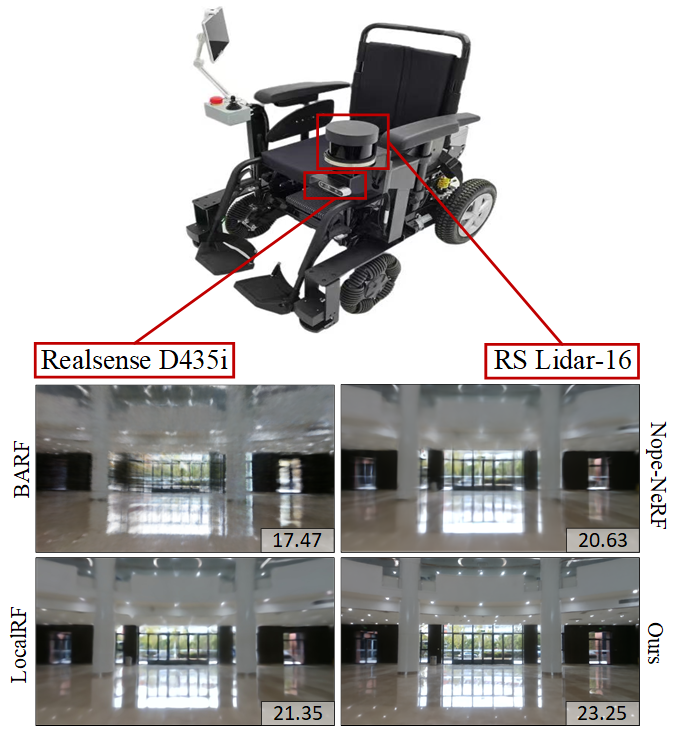}
    \vspace{-0.5cm}
    \caption{
    The image above is a prototype of our robot platform, and below are the qualitative results of our method on our own real-world datasets with PSNR value. We compare the scene reconstruction results with NoPe-NeRF\cite{nopenerf}, LocalRF~\cite{progressive}. Our method performs better on real-world datasets.}
    \label{fig:dataset1}
    \vspace{-0.5cm}
    
\end{figure}

\begin{figure}
    \centering
    \includegraphics[width=\linewidth]{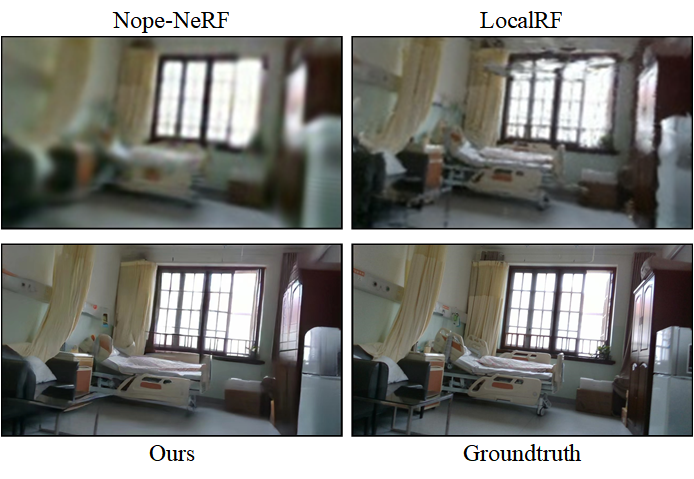}
    \vspace{-0.5cm}
    \caption{
    This image shows the reconstruction results of different methods in a real robot platform within a hospital scenario. We compare the scene reconstruction results with NoPe-NeRF\cite{nopenerf}, LocalRF~\cite{progressive}. Our approach adapts better to data from various scenes, achieving more accurate scene reconstruction and pose estimation.}
    \label{fig:dataset2}
    \vspace{-0.5cm}

\end{figure}

\noindent \textbf{Our Dataset.} An in-house proprietary dataset was collected using an intelligent wheelchair to simulate the application of NeRF in elderly care settings, as shown in Fig.~\ref{fig:dataset1} and Fig.~\ref{fig:dataset2}. We collected datasets and conducted experiments in both laboratory environments and real hospital scenarios. The dataset includes a long-distance travel scene spanning over one hundred meters and consists of over 1k images.
Our intelligent wheelchair features two sensors: the Realsense D435i, an RGBD camera capturing only the RGB images, and the Robosense RS-Lidar-16, aimed at obtaining groundtruth poses with a certified performance at 2 cm level accuracy \cite{nguyen2023slict}.  We evaluate the capabilities of our depth, pose estimation, and scene representation. We use this dataset to validate the performance of our method in real-world robot operations.

 \noindent\textbf{Implementation Details.}
We present our implementation details. We select Adam optimizer\cite{adam} ($\beta=(0.9, 0.999)$) for scene representation and camera tracking optimization. The color loss weighting is $\lambda_1 = 0.25$, $\lambda_2 = 0.1$ for depth, $\lambda_3 = 1.0$ for pose estimation, and $\lambda_4=1.0$, $\lambda_5=1.0$ for optical flow. All training and evaluation experiments are conducted on a single NVIDIA RTX 3090 GPU. Our initial learning rate are $5\times10^{-3}$ for rotations and $5\times10^{-4}$ for translations. The resolution of our local radiance fields is $64^3$. When the camera moves to the bound, we refine the local radiance fields and the camera pose for 600 iterations per frame and upsamle the resolution to $640^3$. After that, we allocate new local radiance fields and repeat this process until all frames are registered.

\begin{table*}[]
\caption{Novel view synthesis on Static Hikes dataset~\cite{progressive} and Proprietary dataset. We use PSNR, SSIM, and LPIPS as our metrics.}
\centering

\begin{tabular}{lllllll|llllll}
\toprule
\multicolumn{1}{l}{\multirow{2}{*}{Methods}} & \multicolumn{6}{c|}{Static Hikes Indoor~\cite{progressive}}                                                 & \multicolumn{6}{c}{Our Dataset}                                                 \\ \cline{2-13} 
 & \multicolumn{1}{c}{PSNR$\uparrow$} & SSIM$\uparrow$ & \multicolumn{1}{c}{LPIPS$\downarrow$} & ATE$\downarrow$ & RPE$_r$$\downarrow$ & RPE$_t$$\downarrow$ & PSNR$\uparrow$ & \multicolumn{1}{c}{SSIM$\uparrow$} & \multicolumn{1}{c}{LPIPS$\downarrow$} & ATE$\downarrow$ & RPE$_r$$\downarrow$ & RPE$_t$$\downarrow$ \\ \midrule

BARF~\cite{barf}                                 & 14.81                    & 0.692 & 0.737                     &  0.632   &   0.545   &   1.791   &   12.38   &      0.56                    &             0.75               &  2.463   &  1.334    &   2.482   \\
NoPe-NeRF~\cite{nopenerf}                             &        16.82                  &   0.74    &       0.62                    &   0.463  &   \textbf{0.443}   &   1.595   &   20.17   &    0.68                      &          0.56                 &  0.970   &   \textbf{0.140}   &   1.459   \\
LocalRF~\cite{progressive}                               & 20.08                    & 0.702 & 0.448                     &   0.231  &    0.846  &   \textbf{1.443}   &   19.63   &   0.649                       &       0.432                    &  0.234   &   0.585   &   0.707   \\
Ours                                  & \textbf{21.65}                    & \textbf{0.709} & \textbf{0.306}                     &   \textbf{0.203}  &   0.941   &    1.752  &   \textbf{22.09}   &    \textbf{0.764}                      &     \textbf{0.277}                      &  \textbf{0.232}   &  0.385   &   \textbf{0.610}   \\ \bottomrule
\end{tabular}
\label{tab:compare}

\end{table*}

\subsection{Experimental Results}
In this section, we evaluate the system in different datasets ranging from indoor scenes to large outdoor scenes on Tanks and Temples~\cite{tanks}, Static Hikes Indoor~\cite{progressive} and our dataset. 

\noindent \textbf{Scene Reconstruction Results.} We use Peak Signal-to-noise Ratio (PSNR), Structural Similarity Index Measurement (SSIM), and Learned Perceptual Image Patch Similarity (LPIPS) to evaluate scene reconstruction results. In Table~\ref{tab:view}, we present the scene reconstruction results on Tanks and Temples datasets. Our method outperforms all the baselines by a large margin, thanks to our incremental scene representation method and efficient tri-plane representation. In Tab.~\ref{tab:tanks}, we compare our method with other scene reconstruction methods with/without COLMAP pose, which shows the proves the advantages of our joint learning system in large-scale scenes. The scene reconstruction results on Static Hikes and our own datasets are shown in Table~\ref{tab:compare}. Our method outperforms other methods in scene representation. The incremental implicit scene representation method significantly enhances the scene representation ability and performs well on large-scale and long-sequence datasets. The qualitative results are shown in Fig.~\ref{fig:reconstruct},\ref{fig:indoor},\ref{fig:dataset1}. In Fig.~\ref{fig:reconstruct}, we present the scene reconstruction results on different sequences Ignatius, Horses, Church in Tanks and Temples dataset~\cite{tanks}. Our method achieves higher-quality image reconstruction results. In Fig.~\ref{fig:indoor}, these images correspond to different timesteps (90, 390, 900) from up to down. We place the PSNR values of each scene in the bottom right corner. In contrast, our incremental scene representation method achieves high-quality scene reconstruction results compared with NoPe-NeRF and LocalRF. In Fig.~\ref{fig:dataset1} and Fig.~\ref{fig:dataset2}, we show the image reconstruction results in our own dataset. Compared with other joint learning methods, our method achieves the most high-fidelity novel view results in both outdoor and indoor scenes. 
\begin{figure}
    \centering
    \includegraphics[width=\linewidth]{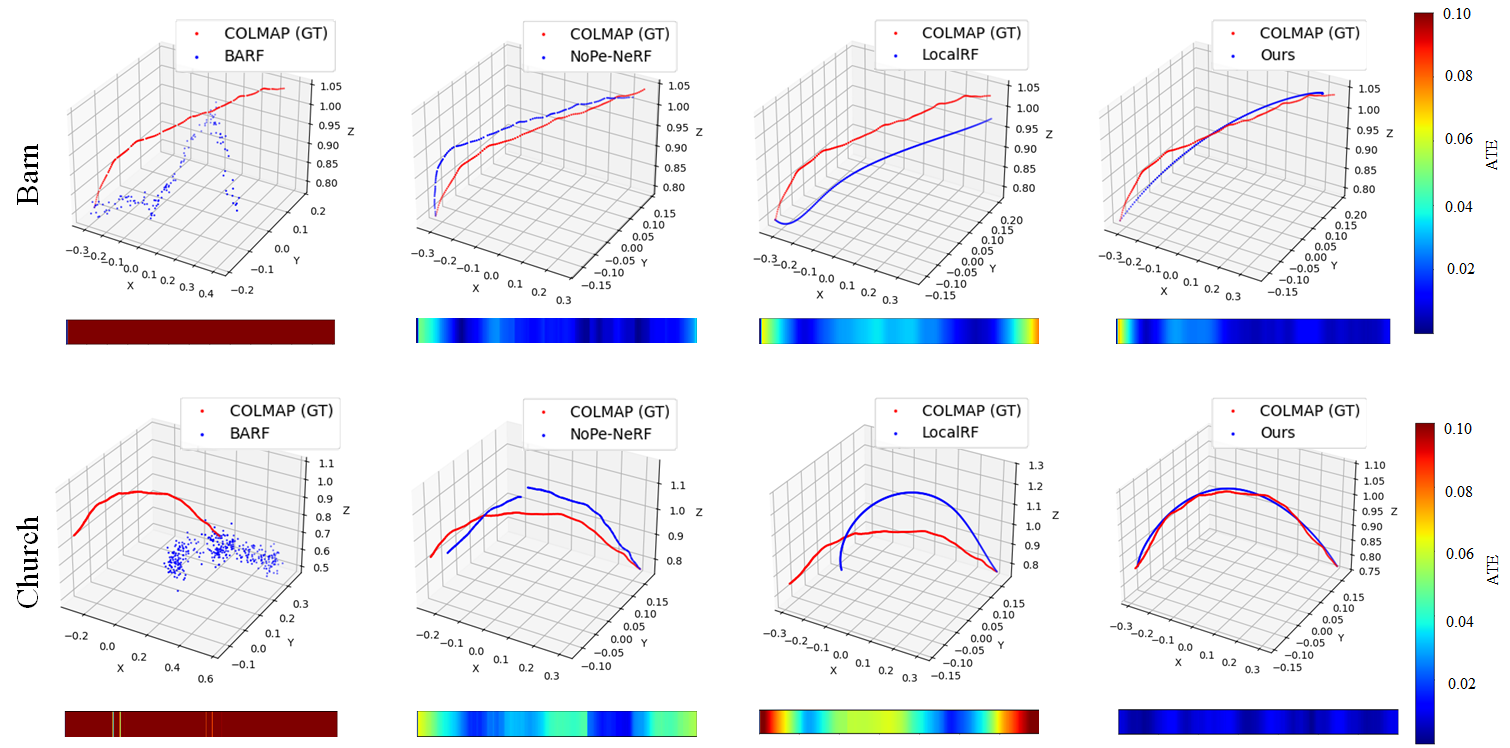}
    
    \caption{Visualization of pose estimation results on Tanks and Temples datasets~\cite{tanks}. We present the pose estimation  results of BARF~\cite{barf}, NoPe-NeRF\cite{nopenerf}, LocalRF~\cite{progressive}, and our method. We visualize the Absolute Trajectory Error ATE  of different methods. The
color bar on the right represents the ATE pose error.}
    \label{fig:pose}

\end{figure}

\begin{figure}
    \centering
    \includegraphics[width=\linewidth]{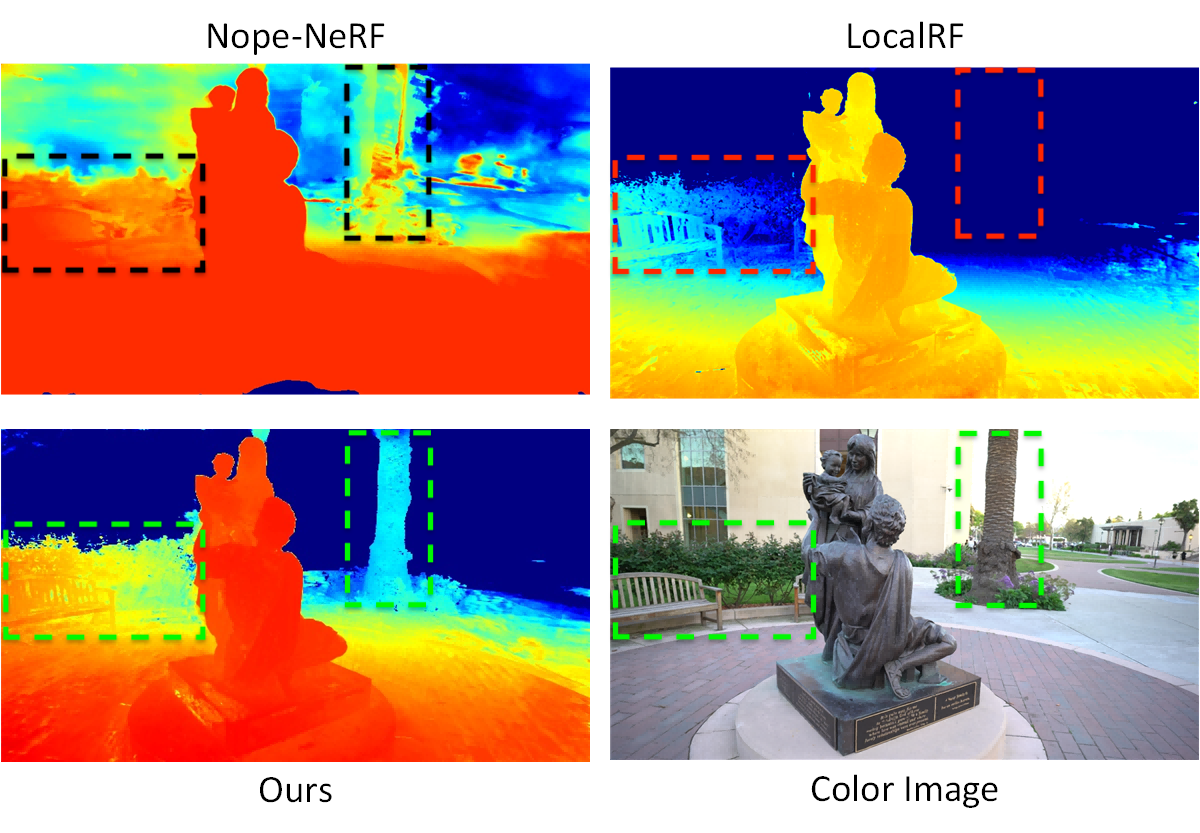}
    
    \caption{Qualitative depth estimation results on the family sequence from the Tanks and Temples dataset~\cite{tanks}. We present the depth estimation results of NoPe-NeRF\cite{nopenerf}, LocalRF~\cite{progressive}, and our method. }
    \label{fig:depth}
    
\end{figure}

\begin{figure}
    \centering
    \includegraphics[width=\linewidth]{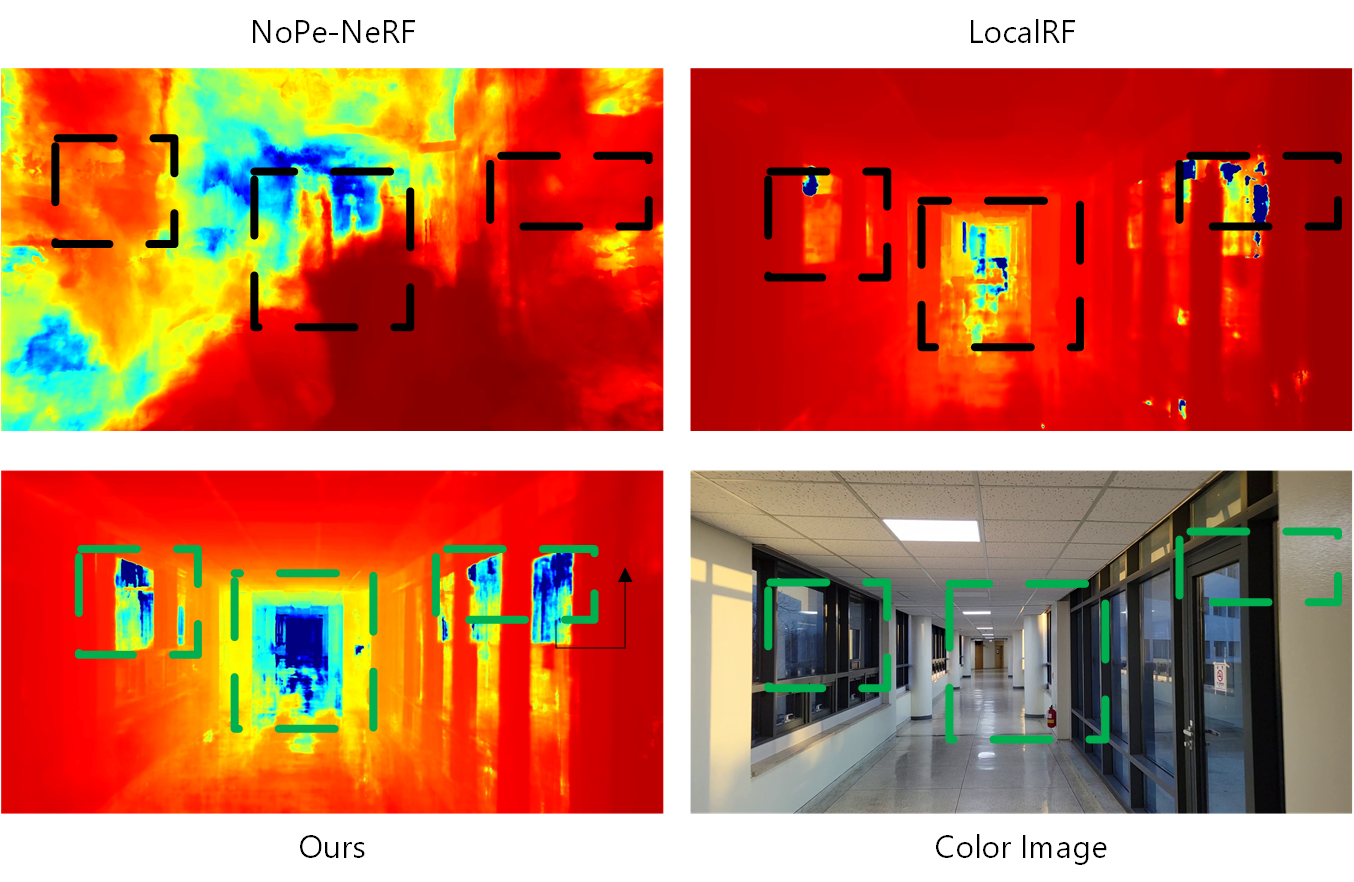}
    
    \caption{Qualitative results of depth estimation on Indoor datasets~\cite{tanks}. We present the depth estimation results of NoPe-NeRF\cite{nopenerf}, LocalRF~\cite{progressive}, and our method. }
    \label{fig:depth3}

\end{figure}

\noindent \textbf{Pose Estimation Results.} The pose estimation results are shown in Table.~\ref{tab:pose} and Table.~\ref{tab:compare}. We use Absolute Trajectory Error (ATE), and Relative Pose Error (RPE) as the pose estimation metrics. Our pose estimation network performs better than other existing methods. We also visualize the ATE error of different methods in Tanks datasets. The FBA method performs re-projection optimization in the high-dimensional feature level, which avoids various disturbances and noise present at the image level. The FBA network enhances the accuracy and robustness of the pose estimation in large-scale scenes. 

\noindent \textbf{Depth Estimation Results.} We present the depth estimation results in Fig.~\ref{fig:depth},\ref{fig:depth3}. In Fig.~\ref{fig:depth}, compared to our method, a tree is missing from the LocalRF estimated depth, as shown in red blocks, and the depth at both bushes and trees in NoPe-NeRF is inaccurate, as shown in black blocks. Our estimated depth images are more accurate and sharper, as shown in green blocks. We  present quantitative results in Tab.~\ref{tab:depth}. Our depth  estimation performance is better than other NeRF-Based methods thanks to the joint learning methods.
\begin{figure}

    \centering
    \includegraphics[width=\linewidth]{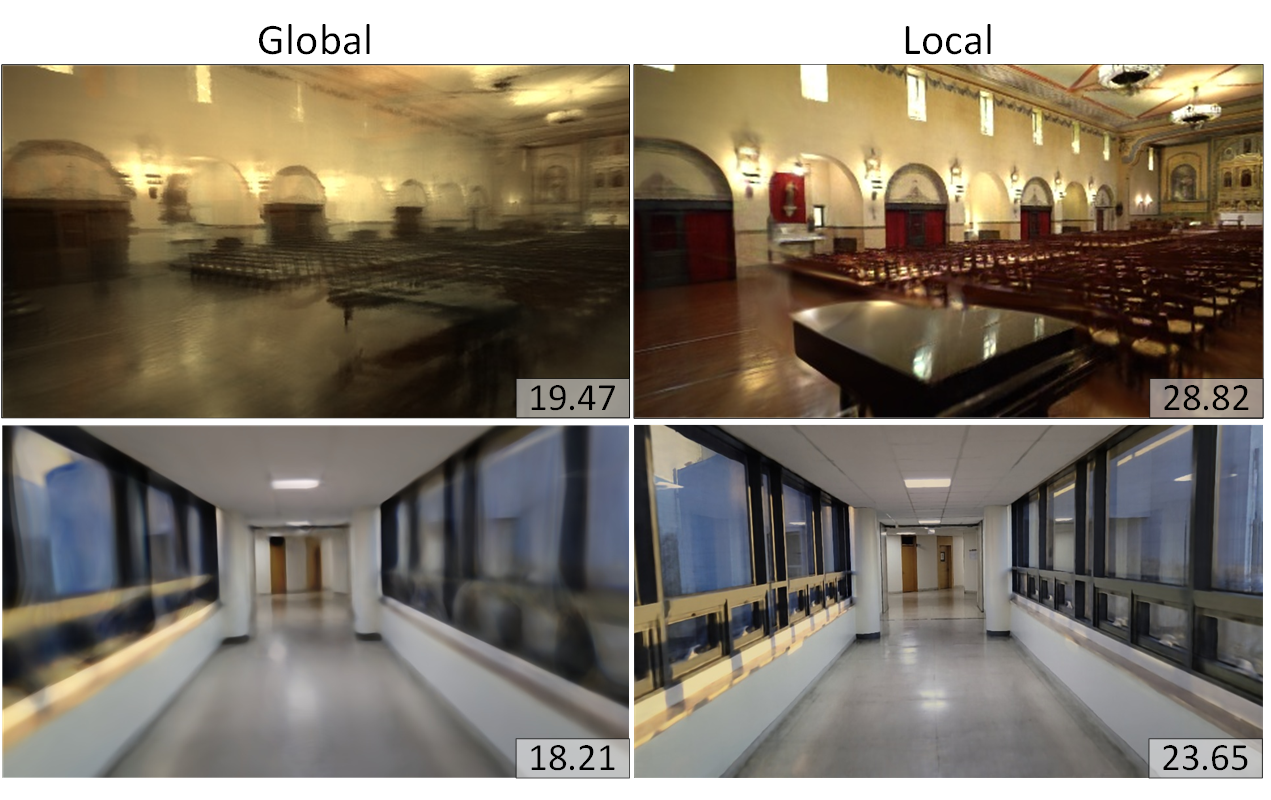}
    
    \caption{Importance of incremental scene representation. We compare the scene reconstruction results with two types of radiance fields model (single, multiple) on Tanks and Temples dataset~\cite{tanks} and Static Hikes dataset~\cite{progressive}. In the first row, we use a single radiance field, which fails in large-scale scenes. In the second row, we use local radiance fields, which improve the representation capability and produce high-fidelity and sharper results.}
    \label{fig:compare}
\end{figure}

\begin{table}[]
\caption{Ablation studies of different modules on Tanks and Temples datasets~\cite{tanks}.}
\setlength{\tabcolsep}{1mm}{
\begin{tabular}{lllllll}
\toprule
\multirow{2}{*}{} & \multicolumn{6}{c}{Tanks and Temples}                                                             \\ \cline{2-7} 
                  & PSNR $\uparrow$          & SSIM $\uparrow$          & LPIPS $\downarrow$        & ATE $\downarrow$           & RPE$_r$ $\downarrow$           & RPE$_t$ $\downarrow$           \\ \midrule
(a) w/o Dep.      & 26.75          & 0.86          & 0.11          & 0.016          & 0.063          & 0.098          \\
(b) w/o ISR       & 26.57          & 0.84          & 0.15          & 0.015          & 0.065          & 0.101          \\
(c) w/o FBA       & 27.15          & 0.87          & 0.10          & 0.016          & 0.068          & 0.103          \\
(d) w COLMAP      & 28.17          & 0.88          & 0.10          & 0.014          & 0.051          & 0.081          \\
Ours              & \textbf{29.04} & \textbf{0.90} & \textbf{0.09} & \textbf{0.012} & \textbf{0.046} & \textbf{0.077} \\ \bottomrule
\end{tabular}}
\label{tab:ablation}

\end{table}

\subsection{Ablation Study}
In this section, we conduct sufficient ablation studies to verify the effectiveness of our designed network. We show the ablation results in Table~\ref{tab:ablation}. (a) we remove the depth estimation and depth prior. (b) is the network without incremental scene representation. (c) is the network without feature-metric bundle adjustment. (d) is our method using pose from COLMAP. In Table (a), we remove the depth loss and the depth network is not jointly optimized with other networks. The results indicate that the depth loss and joint optimization strategy are effective for scene representation and pose estimation.  (b) For the incremental scene representation, we replace it with global optimization. We can see that the incremental scene representation significantly improves the scene representation and pose estimation. We also present the render results between using a single global radiance field and using multiple radiance fields in Fig.~\ref{fig:compare}. This indicates the importance of our incremental scene representation method. (c) We remove the feature-metric bundle adjustment and replace it with traditional pose optimization. We can see that this network improves the accuracy of the pose estimation results. (d) We use the camera pose from COLMAP and fix the camera pose throughout the entire optimization. This demonstrates that our joint optimization method outperforms the original method based on COLMAP.

\section{Conclusion}
In this paper, we propose a novel incremental joint learning network of depth, pose, and implicit scene representation. The proposed FBA network extracts multi-level image features and performs coarse-to-fine bundle adjustment estimation for corresponding pixels and image pose, which can provide accurate pose estimation results both in indoor and outdoor scenes. An incremental scene representation method is designed for large-scale scenes and long-sequence videos. This proposed network can achieve accurate and robust scene representation for arbitrarily long-sequence scene reconstruction. Compared with other joint learning
methods, our method achieves SOTA results in depth estimation, pose estimation, and implicit scene representation in large-scale indoor and outdoor scenes.

\noindent \textbf{Future work}
Currently, our experiments are primarily conducted in relatively large-scale indoor environments. As future work, we plan to extend our approach to even larger, city-scale scenes. To ensure better global consistency in such expansive environments, it is also necessary to incorporate techniques such as loop closure detection into our framework.
\bibliographystyle{IEEEtran}
\bibliography{Bibliography/BIB_xx-xxxx.bib}\ 

\end{document}